# Personalized Deep Learning for Ventricular Arrhythmias Detection on Medical IoT Systems

Zhenge Jia
zhenge.jia@pitt.edu
University of Pittsburgh
Pittsburgh, PA

Zhepeng Wang
zhepeng.wang@pitt.edu
University of Pittsburgh
Pittsburgh, PA

Feng Hong
hongfeng@singularmedical.net
SINGULAR MEDICAL (USA) INC.

Lichuan Ping
lichuanping@singularmedical.net
SINGULAR MEDICAL (USA) INC.

Yiyu Shi
yshi4@nd.edu
University of Notre Dame
Notre Dame, IN

Jingtong Hu
jthu@pitt.edu
University of Pittsburgh
Pittsburgh, PA

## ABSTRACT
Life-threatening ventricular arrhythmias (VA) are the leading cause of sudden cardiac death (SCD), which is the most significant cause of natural death in the US [6]. The implantable cardioverter defibrillator (ICD) is a small device implanted to patients under high risk of SCD as a preventive treatment. The ICD continuously monitors the intracardiac rhythm and delivers shock when detecting the life-threatening VA. Traditional methods detect VA by setting criteria on the detected rhythm. However, those methods suffer from a high inappropriate shock rate and require a regular follow-up to optimize criteria parameters for each ICD recipient. To ameliorate the challenges, we propose the personalized computing framework for deep learning based VA detection on medical IoT systems. The system consists of intracardiac and surface rhythm monitors, and the cloud platform for data uploading, diagnosis, and CNN model personalization. We equip the system with real-time inference on both intracardiac and surface rhythm monitors. To improve the detection accuracy, we enable the monitors to detect VA collaboratively by proposing the cooperative inference. We also introduce the CNN personalization for each patient based on the computing framework to tackle the unlabeled and limited rhythm data problem. When compared with the traditional detection algorithm, the proposed method achieves comparable accuracy on VA rhythm detection and 6.6% reduction in inappropriate shock rate, while the average inference latency is kept at 71ms.

## 1 INTRODUCTION
Sudden cardiac death (SCD) is one of the most significant causes of natural death in the US, which occurs when the heart electrical system malfunctions and leads to 325,000 deaths per year [5]. There are two main causes of SCD, namely, ventricular tachycardia (VT) and ventricular fibrillation (VF), which are the life-threatening ventricular arrhythmias (VA). For people under a great risk of SCD, implantable cardioverter defibrillator (ICD) is served as a preventive treatment. ICD is a small device implanted under the skin with wires inserted in the heart to detect the intracardiac rhythm [30]. This battery-powered device continuously monitors the intracardiac rhythm and is programmed to deliver a shock when VT or VF is detected, and bring the rhythm back to normal.

While ICDs reduce the risk of SCD and increase survival rate, the ICD recipients might experience *inappropriate shocks*, which are the delivery of shock on the rhythm other than VT or VF. *Inappropriate shocks* have been associated with proarrhythmias, intolerable pain, and depression [16]. They have been reported to occur in 12% to 23% ICD recipients and constitute 30% to 50% of all shocks [8, 11, 13]. Current VT/VF detection methods count on a wide variety of criteria on intracardiac rhythm, such as heart rate, the number of intervals to detection (NID), and probability counter [3, 28, 31], where there are hundreds of programmable parameters affecting the detection performance. However, it is complicated to obtain the best parameters setting for each patient as it requires massive clinician experience and frequent manual intervention. Moreover, such optimization cannot be conducted in time since the ICD programming update could only be accomplished every 3-6 months [4].

Recently, deep learning technique has been widely applied on electrocardiogram (ECG) arrhythmias classification using Convolutional Neural Network (CNN) [1, 10, 25]. The CNN-based arrhythmias classification could eliminate the cumbersome criteria selections and parameters setting in traditional arrhythmia detection methods, while achieves decent detection accuracy.

The objective of this paper is to propose a framework for deep learning based VT/VF detection. The framework addresses the following challenges: (1) Existing CNN-based detection models cannot be directly deployed on the implantable devices due to the hardware resources constraints, such as limited memory capacity and computational power; (2) The only intracardiac rhythm sensed by ICDs is the bottleneck for VT/VF detection accuracy improvement. Performing inference on surface and intracardiac rhythm simultaneously could improve the accuracy but the memory and energy overhead would significantly increase; (3) The personalization of detection model for each patient suffers from a long delay and frequent manual intervention on current ICDs platform.

To tackle the above challenges, we propose **P-VA**, a **Personalized** computing framework and architecture for deep learning based **VA** detection. The overview of P-VA is demonstrated in Fig. 1. P-VA consists of two parts: the Detection Nodes and the Cloud. The Detection Nodes, which constitute an implantable node and a wearable node, detect VT/VF on intracardiac electrograms (IEGMs) and surface electrocardiograms (ECG) simultaneously. A CNN is designed and deployed on the both nodes to perform a real-time inference. To further improve the detection accuracy, we propose the *cooperative inference* with a confidence score thresholding policy, which enables two nodes to detect VT/VF collaboratively and significantly reduce the memory and energy overhead by only transmitting the *hardly-decidable* rhythm segments. On the Cloud, we develop an in-time personalization using deep domain adaptation and a policy network to personalize CNNs for the specific patient.

Evaluation results show that our detection nodes could effectively make each inference within 71ms on average. Furthermore, when compared with traditional detection algorithm, our method achieves a comparable VT/VF detection rate and 6.6% reduction in inappropriate shock rate.

The main contributions of this paper are summarized as follows:

- We present P-VA, a personalized computing framework for deep learning based VA detection, to improve the accuracy of VA detection.
- To the best of our knowledge, this paper is the first to deploy a CNN on a resources-constrained and ultra-low power processor with real-time VT/VF detection.
- The cooperative inference reduces the inappropriate shock rate by enabling the implantable and the wearable nodes to detect VT/VF collaboratively.
- We develop an in-time personalization to personalize CNN-based detection model for each specific patient.

The rest of the paper is organized as follows. In Section 2, background and motivation are introduced. Section 3 first provides an overview of the P-VA framework, followed by the construction of CNN-based detection model, the cooperative inference and the model personalization. Evaluation results of the proposed detection methods and its comparisons with the state-of-the-arts are shown in Section 4. Finally, Section 5 concludes the paper.

## 2 BACKGROUND AND MOTIVATION

In this section, we first introduce the background of ICDs and the impedance of improving VT/VF detection accuracy on ICDs. Next we demonstrate the motivation of proposing deep learning based detection and the challenges in application and deployment.

### 2.1 VT/VF Detection in ICDs

The ICD is implanted to deliver defibrillation on VT/VF, which are life-threatening VA leading to SCD. Until now, ICDs are still the best preventive treatment for the population under high risk of SCD. As aforementioned, patients with ICDs, however, might receive inappropriate shocks, which severely affect the patients' quality of life. According to the surveys, 12% to 23% of ICD recipients receive inappropriate shocks in the following 20 months to 11 years after implantation [11, 17, 19, 21]. The study [11] shows that the main cause of inappropriate shock comes from supraventricular arrhythmia misdiagnosed as VT (74.8% of inappropriate shocks). The optimization goal of ICDs is to improve the detection accuracy on VT/VF. However, there are three main obstacles to improve VT/VF detection performance.

**Complex detection criteria and parameter settings.** Current VT/VF detection algorithms count on a combination of various criteria. In the detection programming, there are hundreds of programmable parameters affecting shock delivery decision. The modification of non-nominal parameter settings (i.e., changing the ICD out-of-the-box factory default settings for the recipient) and innovation of new detection criteria are necessary for the improvement of detection accuracy. However, they require extensive clinician proactivity and engagement.

**Long delay in personalized detection.** The criteria parameters are supposed to be optimized to accommodate the unique rhythm feature of each ICD recipient. After implanting an ICD, the recipient should be followed-up every 3-6 months [4]. During the follow-up, the detection criteria parameters would be fine-tuned by a group of doctors based on historical records. However, the detection algorithms could not be updated in time due to the 3-6 months follow-up interval. The in-time personalized detection is unrealistic on the current ICDs platform since it requires massive manual intervention and expertise to fine tune criteria parameters, which severely restrict the personalization frequency.

**Limited rhythm sources.** In the current data sources acquisition, there is an intrinsic bottleneck in detection accuracy improvement, that is, ICDs detect VT/VF purely based on the sensed intracardiac rhythm. The involvement of surface rhythm would improve detection accuracy since there would be more rhythm information in the detection. However, due to the constraints in physical size and lifetime, the computational power and memory capacity of ICDs are severely restricted. The processing unit on ICDs must run with few hundreds of kilobyte of on-chip memory and an active power consumption below the 10 mW mark. Those constraints are the impedance to enable ICDs to detect surface rhythm simultaneously.

### 2.2 Deep Learning in Arrhythmia Detection

Recently, deep learning has been applied to classify arrhythmias on ECGs by using convolutional neural networks (CNNs) and achieved outstanding performance in terms of accuracy [1, 7, 10, 12]. Different from existing detection algorithms, the deep learning based methods eliminates the process of criteria design and optimizations. It significantly reduces the magnitude of cardiological expertise required in the detection parameters fine-tuning. CNN could detect the arrhythmias with high accuracy if the training rhythm data is properly and accurately labeled by doctors.

However, in ICDs scenarios, existing CNN models cannot be directly deployed. The limited memory capacity and real-time inference requirement are the impedance to deploy CNN models on ICDs. Further effort on the CNN architecture design should be applied to fit those hardware constraints. Moreover, only one well-trained CNN as a generalized detection model cannot accommodate all patients due to the unique rhythm features of each patient. The in-time personalization of each specific patient's CNN based detection model remains unsolved. Nevertheless, the improvement of the CNN model detection accuracy is also restricted by the limited rhythm sources. The inference mechanism on both surface and intracardiac rhythm could generate a more accurate prediction.

## 3 PERSONALIZED DEEP LEARNING FOR VA DETECTION

In this section, we first provide an overview of the proposed VT/VF detection framework. Then we introduce the CNN-based detection model. After that, we illustrate the cooperative inference mechanism. Finally, we present the in-time CNN model personalization to accommodate each patient's unique rhythm.

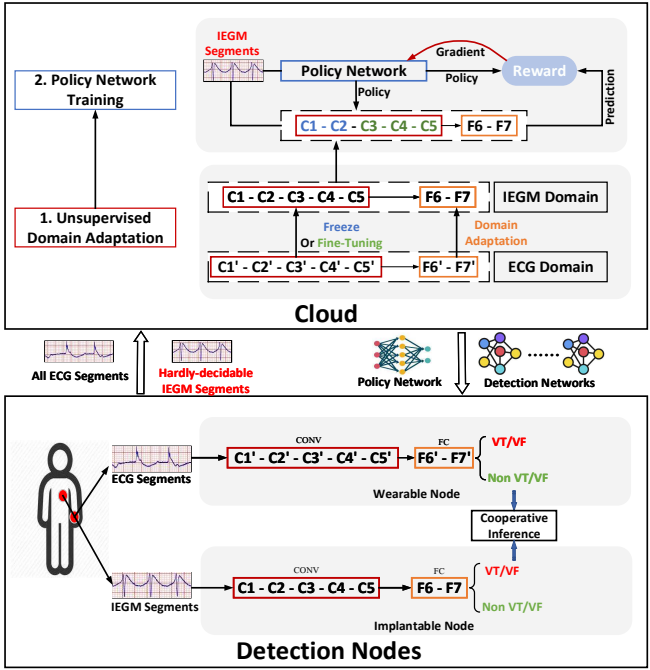

Figure 1: An Overview of P-VA Framework.

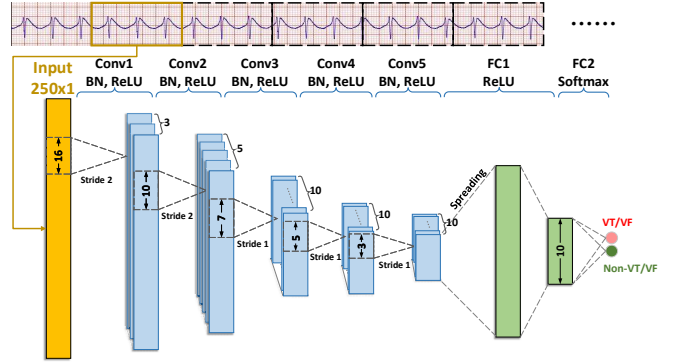

Figure 2: CNN Architecture and Detection Process.

## 3.1 An Overview of P-VA Framework.

In this subsection, we present an overview of the personalized deep learning based VT/VF detection framework. Fig. 1 illustrates our **P-VA** framework, which consists of the Detection Nodes and the Cloud. There are two main functions provided by the framework, the VT/VF detection on Detection Nodes and the detection model personalization on the Cloud.

**Detection Nodes.** The detection nodes consist of the implantable node and the wearable node, and both nodes perform arrhythmias detection on the sensed rhythm. As shown in the Detection Nodes of Fig. 1, the implantable node continuously monitors intracardiac rhythm while the wearable node continuously monitors surface rhythm. Due to the constraints caused by implantation, the implantable node is equipped with an ultra-low power processing unit with few hundreds of KB on-chip memory and an active power consumption below 10 mW. The wearable node is equipped with a more powerful processing unit due to the relatively loose physical-size restrictions and easy access to power supply.

As illustrated in Fig. 1, both nodes conduct VT/VF detection on intracardiac and surface rhythm simultaneously using CNN with the same architecture. The CNN model is designed to satisfy the hardware constraints and perform real-time inference. To further improve the detection accuracy, the cooperative inference is proposed to enable both nodes to detect VT/VF rhythm collaboratively. However, due to the energy constraints, not all IEGMs could be uploaded from the implantable node to the wearable node. Hence a confidence thresholding policy is implemented on the implantable node to identify the limited number of hardly-decidable IEGM segments and trigger the cooperative inference. Moreover, the wearable node would upload the received hardly-decidable IEGMs along with its fully-recorded ECGs to the Cloud for further personalization.

**Cloud.** Once the fully-recorded ECGs and the hardly-decidable IEGMs are uploaded to the Cloud, the fully-recorded ECGs would first be labeled by doctors. However, the uploaded IEGMs could not be accurately labeled due to its non-continuity, and only the CNN on ECG domain could be correctly personalized by retraining.

To tackle the challenge, as illustrated in Fig. 1, we first apply Maximum Mean Discrepancy (MMD) distance to perform unsupervised domain adaptation on the ECG-domain specific CNN to accommodate the IEGM domain [15]. Here, all possible IEGM-domain specific CNNs with different CONV layers being frozen or fine-tuned would be generated as candidate CNNs. In the step 2 on the Cloud in Fig. 1, the policy network is utilized to determine the best-fit CNN from the candidate pool for incoming IEGM segment. Once the domain adaptation and the policy network training completes, the personalized CNNs on ECG and IEGM domains along with the policy network would be propagated back to the wearable and the implantable nodes. Moreover, our framework provides the feasibility that the model could be personalized once the uploaded rhythm is manually labeled by the doctors. Different from traditional upgrading process, our personalization could be completed in a more flexible and timely way, where the patients do not need to wait for 3-6 months to upgrade the detection model.

## 3.2 CNN Detection Model

In this subsection, we introduce the CNN-based VT/VF detection model. Then, we will introduce the mechanism to determine the shockable rhythm on the implantable node.

Since CNN model detects VT/VF on resources-constrained platforms, the network is designed to be relatively small when compared with the existing CNN-based detection models [1, 7, 10, 12]. Fig. 2 illustrates the CNN architecture and the detection process. It consists of five 1D-convolutions (CONV) followed by ReLU function and two fully connected (FC) layers. This CNN-based model requires 26.5 KB to store and the intermediate results during layer-by-layer calculation require at most 4.54 KB.

As demonstrated in Fig. 2, the input of CNN are 2-second (2s) ECG or IEGM segments (i.e., sampling points within the 2s segment) in a time series and the output would be a series of inference results (i.e. VT/VF or non-VT/VF) on each input segment. Based on the series of inference results, on the implantable node, an evaluator would determine the shockable rhythm through a simple but effective decision criteria. This criteria is that the rhythm would be determined as shockable if there are four consecutive VT/VF segments. The detection period on VT/VF is 8 seconds since the detection delay of traditional VT/VF detection algorithm is about 5 to 9 seconds [16]. In other words, if there are four consecutive VT/VF segments in the rhythm, the shock therapy would be triggered.

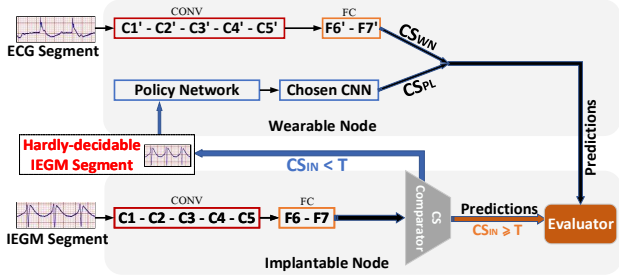

**Figure 3: Cooperative Inference with Confidence Thresholding Policy.**

### 3.3 Cooperative Inference

The CNN-based VT/VF detection performed purely on the implantable node is efficient but it does not achieve the perfect detection due to the limited data sources. A way to improve accuracy is to conduct the inference on intracardiac and surface rhythm through the implantable and the wearable node simultaneously, and to choose the inference result with higher confidence. However, if such detection is performed, both nodes would frequently communicate with each other to send inference results. The transmission overhead and the corresponding energy consumption would greatly reduce the lifetime of the implantable node and increase the response time to the shockable rhythm. Hence, there is a trade-off between accuracy and lifetime on the device. To address the problem, we propose the cooperative inference along with an inference confidence thresholding policy.

The cooperative inference is designed to be simple but efficient to fit the resources constraints. It is based on the concept *Confidence Score (CS)*, which indicates the confidence of the inference on each possible class. As shown in Fig. 3, on the implantable node, its CNN would perform inference on the input IEGM segment. The $CS$ for both classes (i.e., VT/VF and non-VT/VF) on the segment would be calculated and fed into the $CS$ comparator. The $CS$ on the implantable node is defined as follows:

$$CS_{IN} = |P_{IN}(y_{vtvf}|x) - P_{IN}(y_{non-vtvf}|x)|, \quad (1)$$

where $P_{IN}(y|x)$ is the probability of being classified as VT/VF segment ($y_{vtvf}$) or not ($y_{non-vtvf}$) given the input IEGM segment $x$. This probability is obtained by utilizing *Softmax* on the outputs of the last FC layer. Here, a higher $CS_{IN}$ represents a more reliable inference result on the input IEGM segment. When $CS_{IN}$ is low, it reveals the fact that there is a certain level of uncertainty of the inference result, which indicates that the CNN on the implantable could not clearly discriminate the segment. In this case, the CNN on the wearable node should be involved in the VT/VF detection.

A confidence thresholding policy is proposed to utilize a threshold $T$ on $CS$ to determine the participation of the wearable node in VT/VF detection. As shown in Fig. 3, when $CS_{IN} < T$, the corresponding input IEGM segment is defined as the hardly-decidable segment and would be transmitted to the wearable node. On the wearable node, the ECG segment on the same timestamp of the IEGM segment would be the first to feed into CNN and generate the inference result along with the confidence score $CS_{WN}$, and the $CS_{WN}$ on the wearable node is defined as follows:

$$CS_{WN} = |P_{WN}(y_{vtvf}|x) - P_{WN}(y_{non-vtvf}|x)|, \quad (2)$$

where $P_{WN}(y|x)$ is the probability of being classified as VT/VF segment ($y_{vtvf}$) or not ($y_{non-vtvf}$) given the input ECG segment $x$. Moreover, the uploaded hardly-decidable IEGM segment is fed into the policy network (which would be introduced in Section 3.4). Here, the policy network would choose one CNN to make inference on the given IEGM segment and the corresponding $CS$ is defined as $CS_{PL}$. As shown in Fig. 3, by comparing the value of $CS_{PL}$ with $CS_{WN}$, the final inference result (i.e., VT/VF or non-VT/VF) with higher $CS$ value would be transmitted back to the implantable node and fed into the decision evaluator on the implantable node. On the other hand, if $CS_{IN} \geq T$, the inference result generated by CNN on the implantable node would be fed into the evaluator directly without the cooperative inference.

In the cooperative inference, the threshold $T$ affects the chance of data transmission between the implantable node and the wearable node. The determination of the value of $T$ is based on the trade-off among the corresponding performance of accuracy, inference latency and energy consumption. The detailed experimental results and the determination of $T$ will be demonstrated in Section 4.2

### 3.4 In-time CNN Personalization

To cater for the unique features of each patient's rhythm, the CNN-based detection model needs to be personalized with the sensed rhythm. In the P-VA framework, once the data uploaded to the Cloud is accurately labeled, the CNN model personalization would be automatically conducted by fine-tuning CNN with the newly labeled data. This process significantly reduces the degree of manual intervention and the personalization period.

However, in our scenarios, only the arrhythmias on the fully-recorded ECGs could be accurately labeled by doctors since the hardly-decidable IEGM segments are non-continuous and the ventricular arrhythmias on IEGMs could not be correctly diagnosed. Thus, only the ECG-domain specific CNN could be personalized by fine-tuning the CNN on the fully-recorded ECGs.

To obtain the personalized CNN model on IEGM domain, we invoke Deep Adaptation Networks (DAN) proposed in [15] and conduct further optimizations by utilizing the policy network. The core of DAN is to integrate the Maximum Mean Discrepancy (MMD) in the loss function [15]. The invocation of MMD explicitly reduces the domain discrepancy on the FC layers. By minimizing the MMD between the source and the target domain, the mean embedding of distributions across domains can be explicitly matched.

In this work, the source domain is set as labeled ECGs and the target domain is set as unlabeled IEGMs. For the generalized CNN trained on the rhythm from databases, DAN is applied to fine-tune some CONV layers and all FC layers on the ECG domain (i.e., personalize the CNN on ECG domain for the specific patient), and use MMD to tailor the FC layers to fit the un-labelled IEGMs domain (i.e., personalize the CNN on IEGM domain for the specific patient). In this way, the model could be personalized once the uploaded ECG is manually labeled by the doctors. The personalization could be completed with less manual intervention.

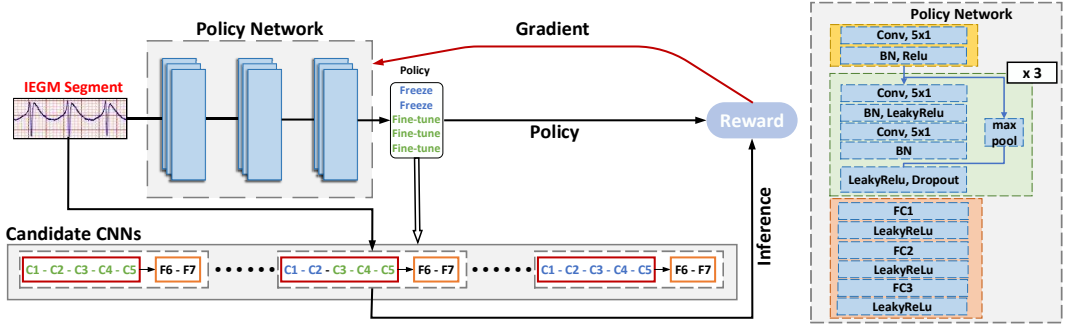

Figure 4: The Illustration of Policy Generation and Training on Policy Network.

DAN employs a global fine-tuning strategy, which fixes the CONV layers to freeze or fine-tune for all input. That is, DAN freezes the low-level CONV layers and fine-tunes the high-level CONV layers on the source domain. It assumes that the low-level CONV layers can learn generic features, while high-level CONV layers are slightly domain-biased. However, the assumption are not met when the target training data is not sufficient or unbalanced. For example, the specific person's IEGM segments might have higher similarity with the ECG segments from rhythm databases, and routing those segments through some the generalized CNN's CONV layers could generate more inference results.

Inspired by the idea from BlockDrop [29], which uses a policy network to dynamically select which layer of a Residual Network to execute during an inference, we utilize the policy network to route each IEGM segment to a specific CNN. As shown in Fig. 4, we first obtain IEGM-domain specific CNNs under all possible freezing or fine-tuning strategy with DAN as the candidate CNNs. Then, the policy network is used to route each incoming IEGM segment for its best-fit IEGM-domain specific CNN. Fig. 4 demonstrates the architecture of the policy network. It consists of one CONV layer, four dense blocks and three FC layers. Inside each dense block, there are two CONV layers with skip connection.

As shown in Fig. 4, there are 32 IEGM-domain specific CNNs (i.e., freeze or fine-tune on each CONV layer out of 5) with the application of DAN. Given an IEGM segment $s$, the policy of deciding which candidate CNN to choose is defined as a 5-dimensional Bernoulli distribution, which takes the form:

$$\pi(\mathbf{a}|\mathbf{s}) = \mathbb{P}[A = \mathbf{a}|S = \mathbf{s}] = \prod_{i=1}^{5} x_i^{a_i}(1-x_i)^{1-a_i}, \quad (3)$$

where $\mathbf{s}$ is the input IEGM segment and $\mathbf{a}$ is the policy that chooses a certain CNN, as shown in Fig. 4. Here, $x$ is the output vector of the policy network after the *Sigmoid* function. The $i$-th element of the vector $x$, $x_i \in [0, 1]$, represents the likelihood of the corresponding CONV layer of the ECG-domain specific CNN being frozen or fine-tuned. The policy $\mathbf{a}$ is also a vector with binary number, which are selected based on $x$, where $a_i = 0$ represents frozen CONV layer $i$ and $a_i = 1$ represents fine-tuned CONV layer $i$.

To encourage the highest accuracy on all segments, as illustrated in Fig. 4, the reward is set based on its prediction correctness of the input segment. This prediction is generated by the IEGM-domain specific CNN selected by the policy network. The reward function associated with the action is defined as

$$R(\mathbf{a}) = \begin{cases} \beta & if\ correct \\ -\beta & otherwise, \end{cases} \quad (4)$$

where $\beta$ is a positive constant. Here, we treat all candidate CNNs equally and reward the correct prediction on the selected CNN with $\beta$. Similarly, incorrect prediction is penalized by $-\beta$. Finally, to obtain the optimal choice of the IEGM-domain specific CNN, we maximize the expected reward associated with the action $ER = \mathbb{E}_\pi[R(\mathbf{a})]$. To maximize the expected value, we utilize the policy gradient to compute the gradients of $J$, which is calculated as follow:

$$\nabla ER = \mathbb{E}_\pi[R(\mathbf{a})\nabla \log \pi(\mathbf{a}|\mathbf{s})] = \mathbb{E}[R(\mathbf{a})\nabla \log \prod_{i=1}^{5} x_i^{a_i}(1-x_i)^{1-a_i}]. \quad (5)$$

Here, we can further optimize the expected gradient to

$$\nabla ER = \mathbb{E}_\pi[R(\mathbf{a})\nabla \sum_{i=1}^{5} \log[x_i a_i + (1-x_i)(1-a_i)]], \quad (6)$$

where the value of $a_i$ is either 0 or 1. Based on the calculated expected gradient on the policy network, we could generate the choice of which IEGM-domain specific CNNs to be applied to the incoming IEGM segment.

After that, the fine-tuned ECG-domain specific CNN, the policy network, and all candidate IEGM-domain specific CNNs would be propagated back from the Cloud to the wearable node. Due to the hardware constraints, the policy network and all candidate CNNs could only be deployed on the wearable node. One IEGM-domain CNN which achieves the highest accuracy during policy network training would be transmitted from the wearable node and deployed on the implantable node.

## 4 EVALUATIONS

We evaluate the performance of the proposed CNN-based VT/VF detection method with a series of experiments. In this section,

Table 1: Database Summary.

| Database | # of Recordings | Rhythm Source | Sampling Rate (Hz) | Recording Duration (min) |
|---|---|---|---|---|
| MITDB | 48 | Lead I | 360 | 30 |
| VFDB | 22 | Lead I | 250 | 30 |
| CUDB | 35 | Lead I | 250 | 8 |
| AAEL | 304 | Lead I, RVA-Bi | 1,000 | 0.5-5 |

Table 2: Training and Testing Dataset for Generalized CNN Training and Personalization.

| Dataset | # of VT/VF Segments | # of Non-VT/VF Segments | Data Source |
|---|---|---|---|
| Generalization-TrainSet | 6,093 | 25,524 | MITDB, CUDB, VFDB |
| Personalization-G1 | 780 | 2,274 | AAEL |
| Personalization-G2 | 781 | 2,275 | AAEL |
| Personalization-G3 | 781 | 2,275 | AAEL |

we first illustrate the experiments setup and then introduce the evaluation results.

### 4.1 Experiments Setup

*4.1.1 Implementations.* In our experiments, the board Apollo3 Blue [27] serves as the implantable node and a Raspberry Pi 3B+ [26] serves as the wearable node. The size of our CNN model is 26.2 KB and requires at most 4.54 KB for intermediate data storage. Such storage requirements could be met by the Apollo3 Blue. Moreover, the total size of the 32 candidate CNN models is 838.4 KB, which is not a large memory overhead for the wearable node.

We adopt PyTorch for the generalized CNN model and the policy network training. For the generalized CNN, we set learning rate to $1e-4$ and use a batch size of 64 during training. We use stochastic gradient descent (SGD) with 0.9 momentum and the loss function is cross-entropy loss function. The total number of epochs is 100. For the policy network, we set learning rate to $1e-4$ and use a batch size of 4 during training. We use SGD with 0.9 momentum and the loss function is cross-entropy loss function. The total number of epochs is 50. All those training experiments run on the PC as the Cloud with 8 core of Intel Xeon E5-2620 v5 CPU, 512 GB memory, and an NVIDIA GeForce GTX 2080Ti GPU.

*4.1.2 Data Preprocessing.* The data used in the experiments is obtained from four databases, namely the MIT-BIH arrhythmia database (MITDB) [18] [23], the MIT-BIH malignant ventricular arrhythmia database (VFDB) [9] [22], the Creighton Univeristy ventricular tachyarrhythmia database (CUDB) [20] [24], and the Ann Arbor Electrogram Libraries (AAEL) [2]. The attributes of the recordings acquired from the four databases are summarized in Table 1. As demonstrated in Table 1, the databases MITDB, CUDB and VFDB only contain the ECG recordings (from Lead I) whereas the recordings in the AAEL are recorded as ECG (from lead I) and IEGM (i.e., from the lead placed in right ventricle, named RVA-Bi) simultaneously. Moreover, the sampling rate of the four databases is different. To ensure the standardization across the databases and reduce inference time, all recordings are downsampled to 125Hz.

The ECG recordings in the MITDB, CUDB and VFDB are utilized to train the CNN on ECG domain as the generalized CNN model. The recordings from those three databases are segmented into 2s non-overlapping segments and each segment is labeled as VT/VF or non-VT/VF using ground truth annotations provided in those three databases. The 2s segments from those three databases constitute the training set for the generalized CNN model. As shown in Table 2, there are 6,093 VT/VF segments and 25,524 non-VT/VF segments in the training set, denoted as Generalization-TrainSet.

The ECG and IEGM recordings in the AAEL are utilized to perform personalization on generalized CNN model and to evaluate detection performance. The data preprocessing procedure for the AAEL is *recording level→ event level→ segment level*. On the recording level, there are 304 recordings as demonstrated in Table 1. Each recording consists of two channels of data from lead I and RVA-Bi. Then, the recording periods diagnosed as VT and VF in the AAEL are labeled as VT/VF events while other are labeled as non-VT/VF events. The reason for the existence of an event level is that the shock therapy is supposed to be delivered during a VT/VF event and the detection accuracy should be calculated based on the correct detection on the events. After labeling and participating in 304 recordings, there are 198 VT/VF and 273 non-VT/VF events in total. Each event is then segmented into non-overlapping 2s segments labeled with VT/VF or non-VT/VF. There are 2,342 VT/VF segments and 6,824 non-VT/VF segments. For the segments in a time series from the same event, we participate those segments into three equally partitioned set and place those set into three groups, denoted as Personalization-G1, G2, and G3. The detailed statistics of each group are demonstrated in Table 2. The rationale behind such grouping is that the segments (i.e., the patient's rhythm from G1 and G2) would be utilized to personalize the model first. The data from the same patient (i.e., the same patient's rhythm from G3) should be applied to evaluate the performance of the personalized CNN model.

*4.1.3 Our Methods and Baseline Methods.* We first obtain the generalized CNN trained on the ECG segments from Generalization-TrainSet. Next, we utilize Personalization-G1 and G2 to personalize the generalized CNN model and utilize Personalization-G3 to evaluate performance. We denote the detection model which is personalized with Personalization-G1 and G2 as CNN-2G. The detection model which is personalized with only Personalization-G1 is denoted as CNN-1G. The generalized CNN which is directly deployed on both nodes are defined as CNN-0G. The cooperative inference and the policy network are activated during the inference. We further implement the CNN-2G without policy network involved. This detection model is used to evaluate the effect of the policy network to detection accuracy. This detection method is denoted as CNN-NoPolicy.

Table 3: Performance metrics for baseline methods and CNN with cooperative inference and policy network.

| Methods | F1 Score | Se/Sp | BAC/Acc | PPV/NPV |
|---|---|---|---|---|
| Classic | .925 | .970/.908 | .939/.934 | .885/.976 |
| SVM-ML | .883 | .874/.923 | .898/.902 | .892/.910 |
| CNN-DL | .901 | .944/.890 | .917/.913 | .862/.957 |
| CNN-NoPolicy | .953 | .970/.952 | .961/.960 | .937/.977 |
| CNN-0G | .831 | .909/.799 | .854/.845 | .766/.924 |
| CNN-1G | .950 | .960/.956 | .958/.958 | .941/.970 |
| CNN-2G | **.975** | **.984/.974** | **.980/.979** | **.965/.989** |

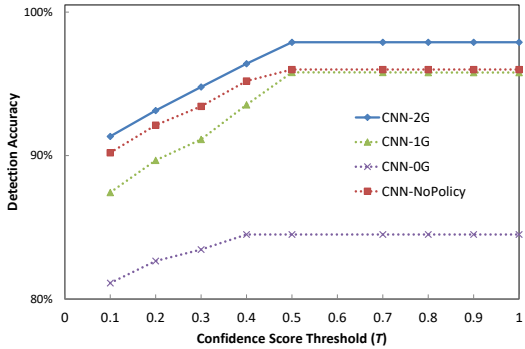

Figure 5: Detection Accuracy vs Confidence Score Threshold.

For the baseline methods, we first simulate a VT/VF discrimination method used in single-chamber ICDs [30], denoted as Classic. In the Classic, the VT zone is set to be 160-200 bpm. The VF zone is set to be the heart rate faster than 200 bpm. The detection window is set as 10 most recent ventricular R-R intervals. The event is determined as the shockable event if the Classic methods decides to deliver the shock. The heart rate boundary of VT/VF zone is also fine-tuned for each testing patient to simulate the intervention such that the best discrimination performance could be achieved. We then implement an existing machine learning based detection method using support vector machine (SVM) [14], denoted as SVM-ML. The features extracted in SVM-ML are *Count2* and *Leakage*. The training set is Personalization-G1 and G2, and testing set is Personalization-G3. We do not implement SVM-ML on the implantable node due to its complex feature extraction process, which consumes extensive computational resources and cannot be completed within 1 second for 2s segment input. Existing machine learning based detection methods are not designed to fit real-time requirements. We also implement a conventional deep learning detection method using CNN. We utilize the IEGM segments from Personalization-G1 and G2 to train the generalized CNN model. The detection performance is evaluated by Personalization-G3 on the implantable node. This detection is denoted as CNN-DL.

## 4.2 Performance Assessment

In this section, we evaluate the performance of CNN-based detection on the P-VA framework with the performance metrics in terms of detection accuracy, latency and energy consumption.

*4.2.1 Detection Accuracy.* The detection accuracy performance is first reported in terms of sensitivity (Se), specificity (Sp), positive and negative predictive accuracy (PPV and NPV), total accuracy (Acc), balanced accuracy (BAC), and F1 score. Those statistics are calculated based on the resulted event detections.

We conduct performance evaluations using the aforementioned detection methods. As shown in Table 3, when compared with the classic VT/VF discrimination algorithm, CNN-DL achieves a slight decrease on performance in terms of VA detection (Se) and inappropriate shock rate (Sp). When compared with SVM-ML, CNN-DL achieves a 1.1% increment from a baseline of 90.2% in Acc and a 6.9% increment on VT/VF event detection rate represented by Se. In other words, the performance of a single CNN on IEGM domain shows that the deep learning based VT/VF detection could achieve decent detection accuracy.

With the assistance of the P-VA framework, the cooperative inference enables CNN model to achieve higher detection accuracy. In the experiment, the confidence score threshold $T$ is set to be 0.5 due to the fact that the accuracy becomes relatively stable when $T > 0.4$. We will demonstrate the accuracy trend along with $T$ in Fig. 5. Here, CNN-2G could achieve the accurate detection, specifically 97.9% accuracy on all events. When compared with the classic detection algorithm, CNN-2G improves by 1.4% on VT/VF event detection accuracy represented by Se and 6.6% on non-VT/VF event detection accuracy represented by Sp. CNN-2G also achieves the best performance in terms of F1 score among all methods. In addition, the personalization process is shown to be necessary for the CNN-based model. As shown in Table 3, CNN-1G experiences a slight degradation in F1 score when compared with CNN-2G. CNN-0G has the worst detection performance among all methods, which we believe is due to the fact that it has no personalization done on the CNN models, so that both node could only make the inference based on the learned features from other patients in rhythm database (i.e., Generalization-TrainSet). Moreover, when compared with CNN-2G, the performance of CNN-NoPolicy illustrates that the policy network could further improve detection accuracy by selecting the best-fit IEGM-domain CNN.

We further investigate the effect of the confidence score threshold $T$ on the detection accuracy. The line plot in Fig. 5 demonstrates the accuracy achieved by CNN-0G, CNN-1G, CNN-2G, and CNN-NoPolicy. These four CNN-based detection methods show to follow the same trend: the accuracy become relatively stable when $T > 0.4$ and the cooperative inference could be more accurate for a larger $T$. It indicates that the appropriate setting of threshold $T$ could achieve the higher detection accuracy through cooperative inference between the wearable and the implantable node.

*4.2.2 Detection Delay.* The experiments aim at assessing the inference latency on the implantable node. On the implantable node,

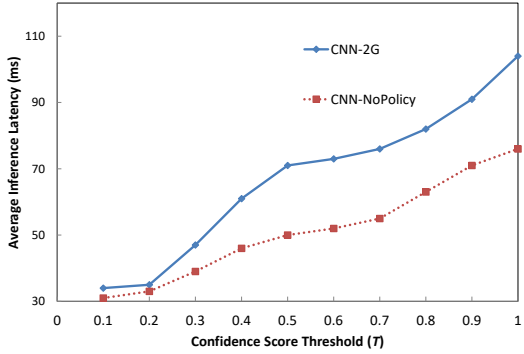

Figure 6: Detection Latency vs Confidence Score Threshold.

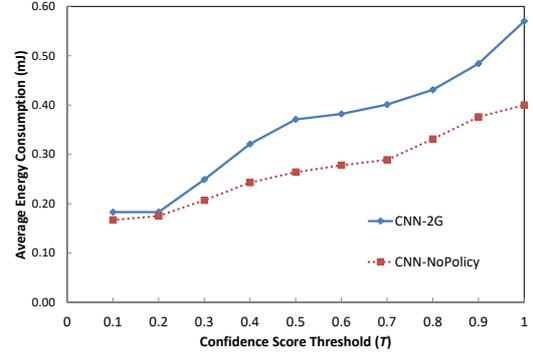

Figure 8: Average Energy Consumption of Inference vs Confidence Score Threshold.

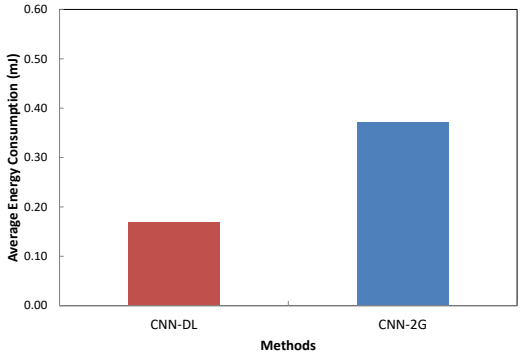

Figure 7: Average Energy Consumption of Inference over One Segment for CNN-DL and CNN-2G.

inference on a 2s IEGM segment could be completed within 31ms without cooperative inference. Here, we do not consider the latency caused by data collection on the implantable node. Data collection is the process of sensing heart rhythm and sending the sensed data to processing unit. Since our focus is on the calculation in inference, we do not consider the latency caused by data collection for all methods. With cooperative inference involved, the actual inference latency on the implantable node should include the transmission delay and the inference latency caused by the wearable node. Here, we investigate the effect of the confidence score threshold $T$ to the inference latency on the implantable node. The line plot in Fig. 6 illustrates the changes of average inference latency on one IEGM segment with an increasing $T$. The performance of two methods, CNN-2G and CNN-NoPolicy, is evaluated in this experiment. Here, both methods show the same trend: the inference latency would increase as the $T$ increases. It is straightforward since the implantable node would be more likely to upload its IEGM segment for cooperative inference as $T$ increases to 1. Moreover, the average inference time of CNN-2G is longer than that of CNN-NoPolicy because the policy network also takes time to execute on the wearable node. One interesting observation is that there is a plateau in the inference latency for both methods, where the $T$ ranges from 0.5 to 0.7. Therefore, $T \subseteq [0.5, 0.7]$ is a suitable value in terms of accuracy and latency.

*4.2.3 Energy Consumption.* In this subsection, we compare average energy consumption of detection for various methods on the implantable node. We also evaluate the effect of the confidence score threshold $T$ on the energy consumption for the CNN-based methods. The energy consumption of the implantable node comes from several operations, including inference and data transmission.

Fig. 7 demonstrates the average energy consumption of inference on a 2s segment for methods CNN-DL and CNN-2G (where $T$ = 0.5) deployed on the implantable node. The performance indicates that the deep learning based methods could achieve a high energy efficiency. As shown in Fig. 7, the energy consumption of CNN-DL is less than that of CNN-2G since the data transmission between the implantable and the wearable node consumes extra energy.

Fig. 8 illustrates the changes of the average energy consumption of inference over a 2s segment along with the confidence score threshold $T$ on the implantable node. The performances of two methods, CNN-2G and CNN-NoPolicy, are evaluated in the experiments. As shown in Fig. 8, both methods would consume more energy with an increasing $T$ since there is a higher probability for the implantable node to transmit data to the wearable node. And CNN-2G would consume more energy because the implantable node should wait for a longer time to receive the results from the wearable node during the cooperative inference.

## 5 CONCLUSION

In this paper, we propose a personalized computing framework for deep learning based VA detection. This framework provides more computing capability for the VA detection. The CNN-based detection model is deployed on the implantable and the wearable nodes to perform real-time VT/VF detection. The cooperative inference is proposed to further improve the detection accuracy by performing the inference based on intracardiac and surface rhythm. On the Cloud, the domain adaptation and the policy network are utilized to perform an in-time CNN model personalization. The proposed deep learning based detection method achieves comparable accuracy on VA detection and 6.6% reduction in inappropriate shock rate when compared with the classical VT/VF discrimination algorithm, while the average inference latency is kept at 71 ms.